\documentclass[]{ceurart}

\usepackage{subcaption}
\usepackage{paralist}

\begin{document}

\copyrightyear{2023}
\copyrightclause{Copyright for this paper by its authors.
  Use permitted under Creative Commons License Attribution 4.0
  International (CC BY 4.0).}

\conference{SEMANTICS 2023 EU: 19th International Conference on Semantic Systems, September 20-22, 2023, Leipzig, Germany}

\title{Developing a Scalable Benchmark for Assessing Large Language Models in Knowledge Graph Engineering}

\author[1,2,3]{Lars-Peter Meyer}[orcid=0000-0001-5260-5181, email=lpmeyer@infai.org]\cormark[1]
\author[1,2,3]{Johannes Frey}[orcid=0000-0003-3127-0815]\fnmark[1]
\author[1,2]{Kurt Junghanns}[orcid=0000-0003-1337-2770]\fnmark[1]
\author[1]{Felix Brei}[orcid=0009-0008-5245-6655]\fnmark[1]
\author[1]{Kirill Bulert}[orcid=0000-0002-1459-3754]
\author[1,2]{Sabine Gründer-Fahrer}[orcid=0000-0003-0054-5003]
\author[1,2]{Michael Martin}[orcid=0000-0003-0762-8688]

\address[1]{Institute for Applied Informatics, Goerdelerring 9, 04109 Leipzig, Germany, \url{https://infai.org}}
\address[2]{Agile Knowledge Engineering and Semantic Web (AKSW), \url{https://aksw.org}}
\address[3]{Leipzig University, Institute for Informatics, Germany, \url{https://www.uni-leipzig.de}}

\cortext[1]{Corresponding author.}
\fntext[1]{These authors contributed equally.}

\begin{abstract}
  As the field of Large Language Models (LLMs) evolves at an accelerated pace, the critical need to assess and monitor their performance emerges. We introduce a benchmarking framework focused on knowledge graph engineering (KGE) accompanied by three challenges addressing syntax and error correction, facts extraction and dataset generation. We show that while being a useful tool, LLMs are yet unfit to assist in knowledge graph generation with zero-shot prompting. Consequently, our \emph{LLM-KG-Bench} framework provides automatic evaluation and storage of LLM responses as well as statistical data and visualization tools to support tracking of prompt engineering and model performance.
\end{abstract}


\maketitle

\section{Introduction}

Large Language Models (LLMs) hold the potential to change the way how we interact with data and technology.
Especially models like GPT-3 and GPT-4 have shown proficient capabilities in solving textual assignments \cite{openai2023gpt4} 
and spawned a wave of subsequent models and the field of \emph{prompt engineering}.

But the fast evolution and rapidly growing landscape of different LLMs make it challenging to keep track of their individual capabilities and to choose the best model and best prompt for the job.
There exist efforts on generic LLM benchmarks (e.g. \cite{Srivastava2022ImitationGameQuantifying}).
However, despite these advancements, the application and (automated) assessment of LLMs in the context of knowledge graph engineering (KGE) and the Semantic Web is still a highly under-explored area.
In response to this gap, this paper proposes a first LLM KGE benchmarking framework \textit{LLM-KG-Bench}\footnote{Repository: \url{https://github.com/AKSW/LLM-KG-Bench} or \href{https://doi.org/10.5281/zenodo.8251944}{doi:10.5281/zenodo.8251944}\label{fn:LLM-KG-Bench-Repo}} that follows our vision of an automated and continuous evaluation platform for different tasks in KGE scenarios. A test of the framework is presented by comparing three LLMs for three exemplary KGE tasks.

\section{Related Work}

The utilization of an LLM in the semantic web domain benefits from its capability to handle RDF-related syntaxes such as JSON-LD, Turtle and SPARQL.
A comprehensive amalgamation of LLMs and knowledge graphs (KGs) is described in \emph{Dagstuhl Seminar} \cite{Groth2023Dagstuhl} and \cite{pan2023unifying}. The \emph{Knowledge Base Construction from Pre-trained Language Models (LM-KBC) Challenge}\footnote{Website: \url{https://lm-kbc.github.io/challenge2023/}} emphasises the relevance of this combination.

The basis of this study is \cite{ours}, where ChatGPT's use in knowledge graph engineering is assessed. Impressive capabilities were revealed, suggesting two conclusions:
Firstly, such studies offer insight into LLMs' potential and limitations, aiding knowledge graph engineers.
Secondly, comparing different LLMs can lead to superior results by addressing inherent model issues.

Recognizing the potential of Large Language Models (LLMs) in knowledge graph engineering, it's vital to evaluate their performance across diverse tasks.
Google's \emph{Beyond the Imitation Game (BIG-bench) Benchmark}\footnote{Repository: \url{https://github.com/google/BIG-bench}\label{fn:BigBenchRepo}}\cite{Srivastava2022ImitationGameQuantifying} and the Large Model Systems (LMSys) leaderboard\footnote{Blogpost: \url{https://lmsys.org/blog/2023-06-22-leaderboard/}\label{fn:LmsysLeaderboard-Blog20230622}} are community efforts that assess the performance of various models with regard to a plethora of tasks.
The \textit{Language Model Evaluation Harness}\footnote{Repository: \url{https://github.com/EleutherAI/lm-evaluation-harness}} offers further testing of generative language models on various evaluation tasks.
However all of them are not perfect for assessing an LLM's use for KGE.
They are missing KGE specific scoring and do not evaluate scores relative to problem size. The size seems to be  relevant for KGE as KGs get quite big in relation to current LLMs context sizes\cite{ours}.
Acknowledging the existing appraoches limitations we introduce the \emph{LLM-KG-Bench} framework.

\section{The LLM-KG-Bench Framework}
Our current (and ongoing) work presented in this paper is comprising the design and implementation of the modular \emph{LLM-KG-Bench} framework\textsuperscript{\ref{fn:LLM-KG-Bench-Repo}} for benchmarking LLMs in the context of knowledge graph engineering.
The main focus is on automated evaluation procedures to allow for many repeated test executions.
The framework supports configurable task sizing, as prior work\cite{ours} suggest the relevance of the LLM's context size for KGE tasks.

As we aim for as much compatibility as possible, especially in the direction of \emph{BIG-bench}\textsuperscript{\ref{fn:BigBenchRepo}}, the \emph{LLM-KG-Bench} framework is organized around \emph{benchmark tasks} and \emph{LLM model connectors}, glued together by some code for execution organisation and result persistence.
\emph{LLM model connectors} encapsulate the connection to a specific LLM and offer the function \verb|generate_text|. With this function a benchmark task can send a prompt to LLM and get its answer. 
\emph{Benchmark tasks} handle the LLM evaluation for a single task.
In the function \verb|evaluate_model| they usually build a prompt or task description for the LLM, hand this task over to a given LLM via an \emph{LLM model connector} and evaluate the given answer.
If necessary the \emph{benchmark task} could send additional prompts to the LLM in the evaluation process.
The evaluation results in score values for the task specific defined score types and additional information.

\begin{figure}
    \centering
    \includegraphics[width=\textwidth]{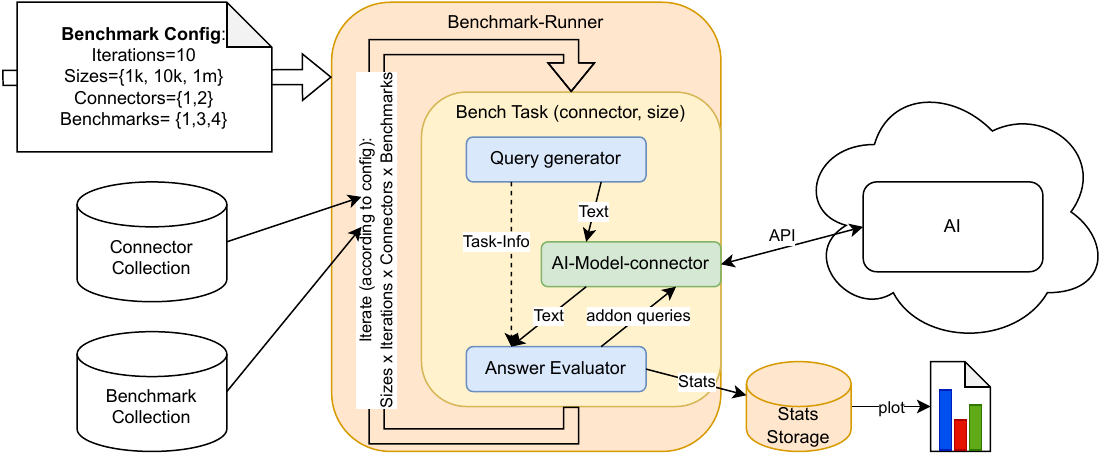}
    \caption{
        Basic \emph{LLM-KG-Bench} framework architecture.
        The Benchmark runner takes a \emph{benchmark configuration} and organizes the repeated execution of \emph{benchmark tasks} with \emph{LLM model connectors} and given size parameters. Results generated get stored and can be visualized.
    }
    \label{fig:architecture}
\end{figure}

Due to \emph{LLM-KG-Bench}'s modularization, as shown in Figure \ref{fig:architecture}, additional benchmark tasks and LLM model connectors can be added by just adding corresponding python class definitions. The framework supports basic result visualization with the help of \emph{seaborn}\footnote{Website: \url{https://seaborn.pydata.org/}}. The plots shown in Figure \ref{fig:three_images} are generated this way.

\section{Initial Evaluation of the Framework with first Tasks}
To test the \emph{LLM-KG-Bench} framework we added a couple of benchmark tasks and evaluated three of the currently highest ranking LLMs at the LLMSYS Chatbot Arena Leaderboard\textsuperscript{\ref{fn:LmsysLeaderboard-Blog20230622}}.
The test setup is detailed in Table \ref{tab:TestsSetup}.

\begin{table}[]
    \caption{Setup used for testing the LLM-KG-Bench framework.}
    \begin{subtable}[h]{0.3\textwidth}
        \begin{tabular}{lllll}
        \toprule
        Model       & Version            \\
        \midrule
        Claude      & claude-1.3-100k    \\
        GPT 3.5 & gpt-3.5-turbo-0613 (4k)        \\
        GPT 4   & gpt-4-0613 (8k)\\
        \bottomrule
        \end{tabular}
        \caption{LLMs evaluated}
        \label{tab:LlmVersions}
    \end{subtable}
    \hfill
    \begin{subtable}[h]{0.6\textwidth}
        \begin{tabular}{llll}
        \toprule
                       & Task a      & Task b      & Task c       \\
        \cmidrule{2-4}
        Repetitions:    & 20 x 1 size & 20 x 1 size & 20 x 8 sizes \\
        plot type:      & F1 measure  & F1 measure  & Mean error   \\
        plot generated: 
            & Figure \ref{fig:TurtleErrorFixingScores}
            & Figure \ref{fig:plotFactsExtract}
            & Figure \ref{fig:TestDataGeneration} \\
        \bottomrule
        \end{tabular}
        \caption{test configuration per task}
        \label{tab:my-table}
    \end{subtable}

    \label{tab:TestsSetup}
\end{table}

\begin{figure}
  \centering
  \begin{subfigure}{0.25\textwidth}
    \centering
    \includegraphics[width=\linewidth]{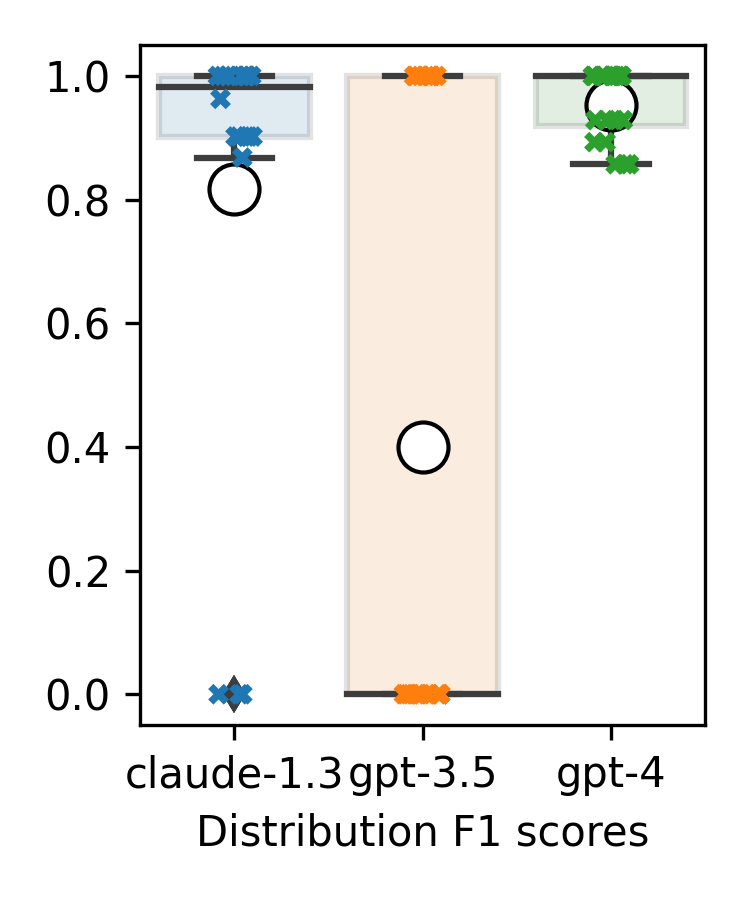}
    \caption{Turtle Fixing}
    \label{fig:TurtleErrorFixingScores}
  \end{subfigure}%
  \hfill
  \begin{subfigure}{0.25\textwidth}
    \centering
    \includegraphics[width=\linewidth]{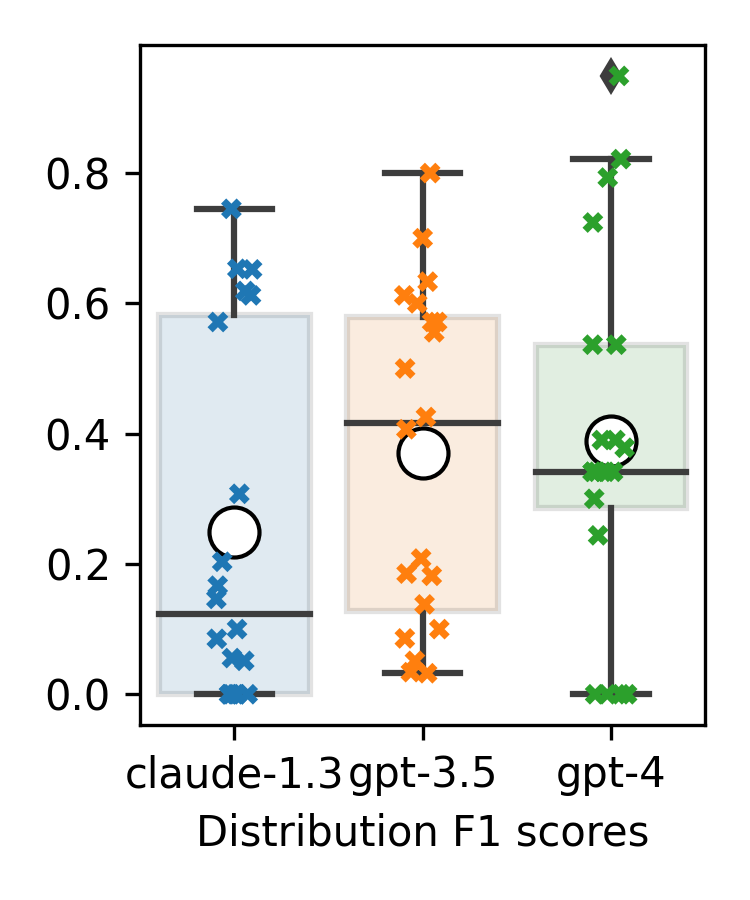}
    \caption{Fact Extraction}
    \label{fig:plotFactsExtract}
  \end{subfigure}%
  \hfill
  \begin{subfigure}{0.5\textwidth}
    \centering
    \includegraphics[width=\linewidth]{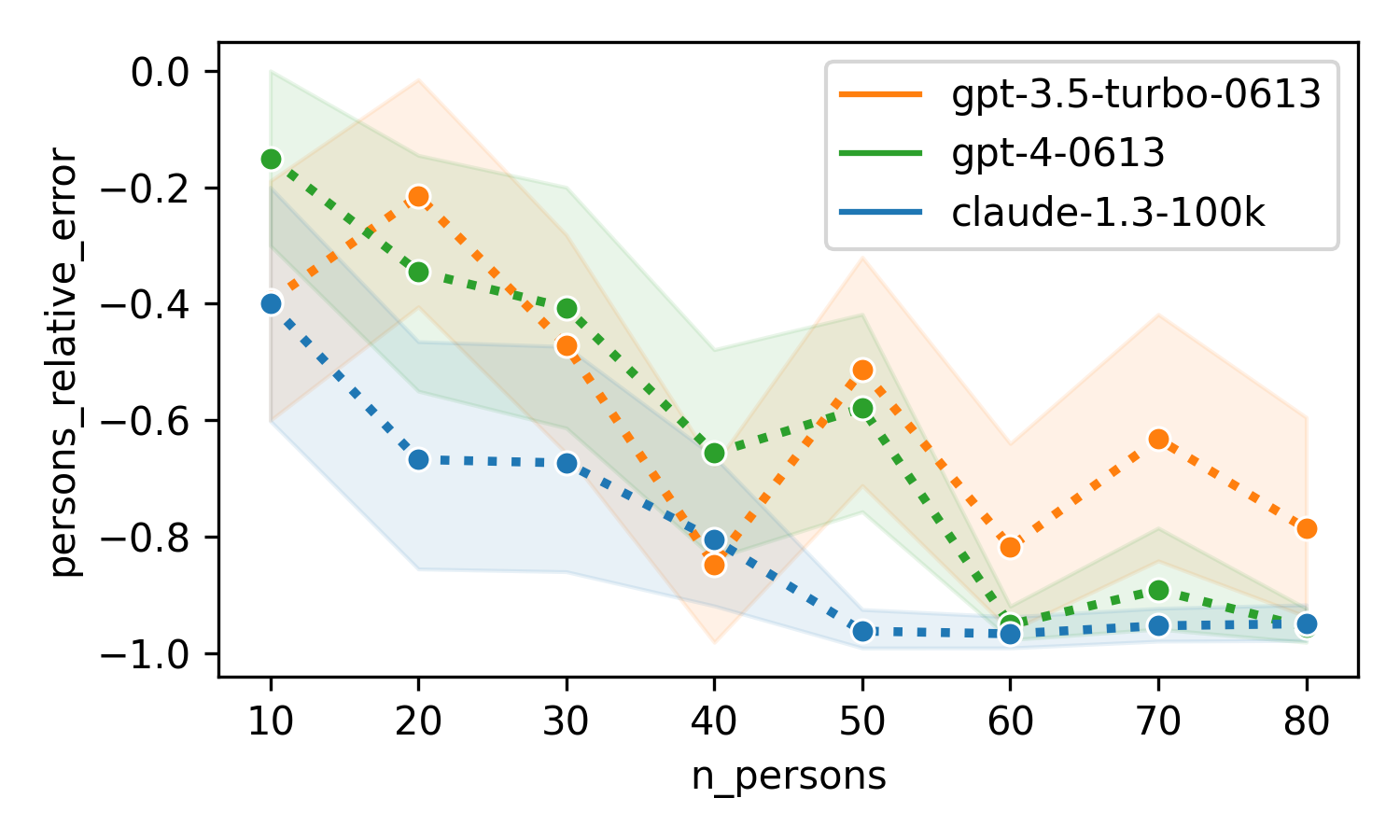}
    \caption{Mean Error Dataset Generation}
    \label{fig:TestDataGeneration}
  \end{subfigure}
  \caption{Subset of metrics from initial tasks. Shown are the F1 scores and mean error of person count}
  \label{fig:three_images} 
\end{figure}

\subparagraph{Task a: Fixing of Errors in Turtle Files:}
\label{sec:TurtleErrorFixing}
Turtle is a common serialization format for knowledge graphs.
By asking the LLMs to fix errors in given manipulated turtle files we test the knowledge of turtle syntax as well as strict adhering to the given task and facts.
One of the scores calculated during evaluation is the F1 measure on parsable normalized triples, comparing LLM's answer with a perfect answer.
A plot on the F1 measure results for this task is shown in Figure \ref{fig:TurtleErrorFixingScores}.
GPT-3.5 often claims that file would be correct and returns no turtle. 
This accounts for the high frequency of zero-value F1 scores.
The answers given by Claude-1.3 and GPT-4 score better.

\subparagraph{Task b: KG Creation from Factsheet Plaintext:}
\label{sec:FactsExtract}
To evaluate knowledge extraction and modelling capabilities, we use a plaintext excerpt of a PDF factsheet.  
The text describes various specifications of a 3D printer in a key-value style, including usual formatting irregularities associated with PDF extraction.
We ask the model to generate a Turtle file, that captures a subset of the information. 
The prompt is engineered very specific with regard to which properties or ontologies have to be used and how IRI identifiers and Literals should be represented.
Subsequently, we can evaluate the quality of a single response using the F1 measure, counting the set of parsable triples that (mis)match or are missing compared to a manually curated reference document.
Fig. \ref{fig:plotFactsExtract} shows that the GPT models outperform Claude in this task. 
While GPT4 has a better mean, due to one very good response, it however replied often with unparseable content, which in turn did not happen for GPT3.5, leading to a slightly better median for that.

\subparagraph{Task c: Synthetic Dataset Generation:}
\label{sec:TestDataGeneration}
Creating example data is an important task and the help of LLMs would be highly appreciated.
We created a basic test for this capability. We ask the LLM to generate some synthetic dataset using well known \verb|foaf:Person| and \verb|foaf:knows| with a varying number of desired objects and links in the final KG.
In the evaluation we used beside other scores the \emph{persons\_relative\_error} indicating the difference between the actual number person objects generated and the number asked for.
This value is normalized to be $=0$ if they match, $>0$ if there are more persons than asked for and $<0$ if there are less persons, with the special case of $-1$ meaning an empty graph.
The results presented in Figure \ref{fig:TestDataGeneration} show a relation between the \emph{persons\_relative\_error} and the problem size, in this case number of person objects to generate.

\section{Conclusion and Future Work}
We showed that there is a need for measuring the knowledge graph engineering capabilities of the rapidly evolving LLMs.
We proposed and describe the novel \emph{LLM-KG-Bench} framework for this task.
A first evaluation of three high ranking LLMs with first benchmarks shows the benefit of the automated evaluation with the new framework.

The \emph{LLM-KG-Bench} framework is prepared to enable dialogs between benchmark tasks and LLMs. It will be interesting to evaluate LLMs capabilities to fix their answers with some feedback like e.g. error codes in improved or additional tasks.
We are looking forward to extending to more LLMs and more benchmark tasks with the help of a bigger community.

\begin{acknowledgments}
  This work was partially supported by grants from the German Federal Ministry for Economic Affairs and Climate Action (BMWK) to the CoyPu project (01MK21007A) and KISS project (01MK22001A) as well as from the German Federal Ministry of Education and Research (BMBF) to the projects StahlDigital (13XP5116B) and KupferDigital (F13XP5119F).
\end{acknowledgments}

\bibliography{ref}

\begin{thebibliography}{5}
\expandafter\ifx\csname natexlab\endcsname\relax\def\natexlab#1{#1}\fi
\providecommand{\url}[1]{\texttt{#1}}
\providecommand{\href}[2]{#2}
\providecommand{\path}[1]{#1}
\providecommand{\DOIprefix}{doi:}
\providecommand{\ArXivprefix}{arXiv:}
\providecommand{\URLprefix}{URL: }
\providecommand{\Pubmedprefix}{pmid:}
\providecommand{\doi}[1]{\href{http://dx.doi.org/#1}{\path{#1}}}
\providecommand{\Pubmed}[1]{\href{pmid:#1}{\path{#1}}}
\providecommand{\bibinfo}[2]{#2}
\ifx\xfnm\relax \def\xfnm[#1]{\unskip,\space#1}\fi
\bibitem[{OpenAI(2023)}]{openai2023gpt4}
\bibinfo{author}{OpenAI}, \bibinfo{title}{Gpt-4 technical report},
  \bibinfo{year}{2023}. \href{http://arxiv.org/abs/2303.08774}{{\tt
  arXiv:2303.08774}}.
\bibitem[{Srivastava et~al.(2023)}]{Srivastava2022ImitationGameQuantifying}
\bibinfo{author}{A.~Srivastava}, et~al.,
\newblock \bibinfo{title}{Beyond the imitation game: Quantifying and
  extrapolating the capabilities of language models},
\newblock \bibinfo{journal}{Transactions on Machine Learning Research}
  (\bibinfo{year}{2023}). \href{http://arxiv.org/abs/2206.04615}{{\tt
  arXiv:2206.04615}}.
\bibitem[{Groth et~al.(2023)Groth, Simperl, van Erp, and
  Vrandečić}]{Groth2023Dagstuhl}
\bibinfo{author}{P.~Groth}, \bibinfo{author}{E.~Simperl},
  \bibinfo{author}{M.~van Erp}, \bibinfo{author}{D.~Vrandečić},
\newblock \bibinfo{title}{Knowledge graphs and their role in the knowledge
  engineering of the 21st century (dagstuhl seminar 22372)}
  (\bibinfo{year}{2023}). \DOIprefix\doi{10.4230/DAGREP.12.9.60}.
\bibitem[{Pan et~al.(2023)Pan, Luo, Wang, Chen, Wang, and Wu}]{pan2023unifying}
\bibinfo{author}{S.~Pan}, \bibinfo{author}{L.~Luo}, \bibinfo{author}{Y.~Wang},
  \bibinfo{author}{C.~Chen}, \bibinfo{author}{J.~Wang},
  \bibinfo{author}{X.~Wu}, \bibinfo{title}{Unifying large language models and
  knowledge graphs: A roadmap}, \bibinfo{year}{2023}.
  \href{http://arxiv.org/abs/2306.08302}{{\tt arXiv:2306.08302}}.
\bibitem[{Meyer et~al.(2023)Meyer, Stadler, Frey, Radtke, Junghanns, Meissner,
  Dziwis, Bulert, and Martin}]{ours}
\bibinfo{author}{L.-P. Meyer}, \bibinfo{author}{C.~Stadler},
  \bibinfo{author}{J.~Frey}, \bibinfo{author}{N.~Radtke},
  \bibinfo{author}{K.~Junghanns}, \bibinfo{author}{R.~Meissner},
  \bibinfo{author}{G.~Dziwis}, \bibinfo{author}{K.~Bulert},
  \bibinfo{author}{M.~Martin}, \bibinfo{title}{Llm-assisted knowledge graph
  engineering: Experiments with chatgpt}, \bibinfo{year}{2023}.
  \href{http://arxiv.org/abs/2307.06917}{{\tt arXiv:2307.06917}},
  \bibinfo{note}{to appear in proceedings of AI-Tomorrow track on Data Week
  2023 in Leipzig}.

\end{thebibliography}

\appendix

\section{Online Resources}

\begin{compactitem}
    \item \emph{LLM-KG-Bench} repository: \url{https://github.com/AKSW/LLM-KG-Bench} \\
    or \href{https://doi.org/10.5281/zenodo.8251944}{doi:10.5281/zenodo.8251944}
    \item experiment data: \url{https://github.com/AKSW/LLM-KG-Bench-Results/tree/main/2023-SEMANTICS_LLM-KGE-Bench-Results} or \href{https://doi.org/10.5281/zenodo.8250646}{doi:10.5281/zenodo.8250646}
\end{compactitem}

\end{document}